\documentclass[runningheads]{llncs}
\usepackage[T1]{fontenc}
\usepackage{graphicx,verbatim}
\usepackage{amsmath}
\usepackage{amssymb}
\usepackage{amsfonts}
\usepackage{amscd}
\usepackage{graphicx}
\usepackage{bm}
\usepackage{color}
\usepackage{xcolor}
\usepackage{authblk}
\usepackage{booktabs}
\usepackage{multirow}
\usepackage{float}
\usepackage{tikz}
\usetikzlibrary{positioning}
\usepackage{url}
\usepackage{pifont}
\usepackage{listings}
\usepackage{breakcites}
\usepackage{orcidlink}
\usepackage{setspace}
\usepackage{latexsym}
\usepackage{caption}
\usepackage{makecell}
\usepackage{array}
\usepackage{wrapfig}
\usepackage{diagbox}
\usepackage{algorithm}
\usepackage{algpseudocode}
\usepackage{textcomp}
\usepackage[T1]{fontenc}
\usepackage{upquote}
\usepackage{dashrule}
\usepackage{subfig}
\usepackage{tabularx}
\usepackage[toc, page]{appendix}
\usepackage[noabbrev,capitalise]{cleveref}
\usepackage{seqsplit}
\usepackage{url} 
\usepackage{lipsum}
 \usepackage{hyperref}

\usetikzlibrary{arrows.meta,arrows,fit}

\definecolor{Gray}{gray}{0.9}

\definecolor{citecolor}{HTML}{0071bc}

\usepackage[toc, page]{appendix}
\usepackage{tabularx}
\usepackage{booktabs}

\newlength\savewidth
  
\usepackage{amsmath,amsfonts,bm}

\renewcommand{\mathbf}{\boldsymbol}

\newcommand{\te}{\texttt}

\makeatletter
\def\Ddots{\mathinner{\mkern1mu\raise\p@
\vbox{\kern7\p@\hbox{.}}\mkern2mu
\raise4\p@\hbox{.}\mkern2mu\raise7\p@\hbox{.}\mkern1mu}}
\makeatother
\newcommand{\blue}[1]{{\color[HTML]{6C8EBF} #1}}
\newcommand{\brown}[1]{{\color[HTML]{80461B} #1}}
\newcommand{\green}[1]{{\color[HTML]{5F8940} #1}}
\definecolor{darkgreen}{rgb}{0.0, 0.5, 0.0}
\definecolor{darkred}{rgb}{0.88, 0.09, 0.12}

\numberwithin{equation}{section}

\usepackage{tcolorbox}
\tcbuselibrary{skins}

\newtcolorbox{userinput}{
    colback=white!10,
    colframe=gray!40,
    fonttitle=\bfseries,
    coltitle=black,
    colbacktitle=gray!20,
    enhanced,
    drop shadow=black!5!white,
    left=8mm,
    right=8mm,
    top=3mm,
    bottom=3mm,
    boxsep=0mm,
    sharp corners=south,
    rounded corners=north,
    title=EMT Prompt:
    
    }

\newtcolorbox{userinputclip}{
    colback=white!10,
    colframe=gray!40,
    fonttitle=\bfseries,
    coltitle=black,
    colbacktitle=gray!20,
    enhanced,
    drop shadow=black!5!white,
    left=8mm,
    right=8mm,
    top=3mm,
    bottom=3mm,
    boxsep=0mm,
    sharp corners=south,
    rounded corners=north,
    title=Prompt:
    
    }

\newtcolorbox{modeloutput}[2]{
    colback=citecolor!5,
    colframe=gray!40,
    fonttitle=\bfseries,
    coltitle=black,
    colbacktitle=gray!20,
    enhanced,
    drop shadow=black!5!white,
    left=8mm,
    right=8mm,
    top=3mm,
    bottom=3mm,
    boxsep=0mm,
    sharp corners=south,
    rounded corners=north,
    title=Label: {#1} \;|\; {\te{#2}}: 
    
    }

\newtcolorbox{modeloutput_withimage}[3][]{
    colback=citecolor!5,
    colframe=gray!40,
    fonttitle=\bfseries,
    coltitle=black,
    colbacktitle=gray!20,
    enhanced,
    drop shadow=black!5!white,
    left=1.5cm, % Adjusted for the image
    right=8mm,
    top=3mm,
    bottom=3mm,
    boxsep=0mm,
    sharp corners=south,
    rounded corners=north,
    title=Label: {#2} \;|\; {\te{#3}},
    overlay={
        \node[anchor=north west, yshift=-4.1mm, xshift=0.4mm] (imgnode) at (frame.north west) {\includegraphics[width=0.7cm]{#1}};
        \draw[gray!50, line width=0.5mm] (imgnode.east |- frame.north) -- (imgnode.east |- frame.south);
        
        \node[anchor=south, font=\footnotesize, text=black, yshift=-1mm, xshift=0mm] at (imgnode.north) {\te{img}};

    }
}

\begin{document}

\title{MedVLM-R1: Incentivizing Medical Reasoning Capability of Vision-Language Models (VLMs) via Reinforcement Learning}

\authorrunning{Pan, Liu et al.}
\titlerunning{MedVLM-R1}

\author{Jiazhen Pan\inst{1,2}$^{*}$, Che Liu\inst{3}$^{*}$, Junde Wu\inst{2}, Fenglin Liu\inst{2}, Jiayuan Zhu\inst{2}, Hongwei Bran Li\inst{4}, Chen Chen\inst{5,6}, Cheng Ouyang\inst{2,6}$^{\dagger}$, Daniel Rueckert\inst{1,6}$^{\dagger}$}
\institute{Chair for AI in Healthcare and Medicine, Technical University of Munich (TUM) and TUM University Hospital, Germany 
\and Department of Engineering Science, University of Oxford, UK
\and Data Science Institute, Imperial College London, UK
\and Massachusetts General Hospital, Harvard Medical School, USA
\and School of Computer Science, University of Sheffield, UK
\and Department of Computing, Imperial College London, UK
\\
\email{jiazhen.pan@tum.de, che.liu21@imperial.ac.uk}
}
 
\maketitle              % typeset the header of the contribution
\let\thefootnote\relax\footnotetext{$^{*}$ Equal contribution}
\let\thefootnote\relax\footnotetext{$^{\dagger}$ Equal advice}

\begin{abstract} 

Reasoning is a critical frontier for advancing medical image analysis, where transparency and trustworthiness play a central role in both clinician trust and regulatory approval. Although Medical Visual Language Models (VLMs) show promise for radiological tasks, most existing VLMs merely produce final answers without revealing the underlying reasoning. To address this gap, we introduce MedVLM-R1, a medical VLM that explicitly generates natural language reasoning to enhance transparency and trustworthiness. Instead of relying on supervised fine-tuning (SFT), which often suffers from overfitting to training distributions and fails to foster genuine reasoning, MedVLM-R1 employs a reinforcement learning framework that incentivizes the model to discover human-interpretable reasoning paths without using any reasoning references. Despite limited training data (600 visual question answering samples) and model parameters (2B), MedVLM-R1 boosts accuracy from 55.11\% to 78.22\% across MRI, CT, and X-ray benchmarks, outperforming larger models trained on over a million samples. It also demonstrates robust domain generalization under out-of-distribution tasks. By unifying medical image analysis with explicit reasoning, MedVLM-R1 marks a pivotal step toward trustworthy and interpretable AI in clinical practice. Inference model is available at: \url{https://huggingface.co/JZPeterPan/MedVLM-R1}.

% \keywords{Medical reasoning  \and Reinforcement learning \and VLMs}

\end{abstract}

\section{Introduction}

Radiological images are fundamental to modern healthcare, with over 8 billion scans performed annually \cite{akhter2023ai}. As diagnostic demand grows, the demand for efficient AI-driven interpretation becomes increasingly acute. Medical Vision-Language Models (VLMs), developed for radiological visual question answering (VQA) in MRI, CT and X-ray images, offer substantial promise in assisting clinicians/patients. Recent advances in general-purpose LLMs/VLMs (e.g., GPT-4o~\cite{hurst2024gpt}, Claude-3.7 Sonnet~\cite{claude-37}) highlight sophisticated reasoning capabilities. However, the medical domain places an especially high premium on explainable decision-making: both clinicians/patients need to understand not just \emph{what} conclusion was reached, but also \emph{why}. Existing medical VLMs often provide only final answers or “quasi-explanations” derived from pre-training pattern matching, which do not necessarily reflect genuine, step-by-step reasoning. Consequently, ensuring interpretability and trustworthiness remains an urgent challenge in real-world clinical settings.

We argue that the limited reasoning capability for existing medical VLM is primarily due to the inherent drawbacks of Supervised Fine-Tuning (SFT) \cite{achiam2023gpt,qwq-32b-preview,chen2022program} which is the most common strategy for adapting large foundation models for specialized medical tasks \cite{hartsock2024vision,lian2024less,chen2024efficiency}. Despite its simplicity, SFT faces two critical challenges: 1) An over-reliance on final-answer supervision often leads to overfitting, shortcut learning, and weaker performance on out-of-distribution (OOD) data -- an issue particularly consequential in high-stake medical scenarios \cite{chu2025sft}.
2) Direct supervision with only final answers provides minimal incentive for cultivating reasoning abilities within VLMs. A possible mitigation is distilling a more capable teacher model's chain-of-thought (CoT) reasoning for SFT \cite{wei2022chain,li2023symbolic}. However, constructing high-quality CoT data is prohibitively expensive to scale in specialized domains like healthcare. As a result, current medical VLMs that rely on SFT often fall short of delivering transparent explanations and robust generalizations when confronted with unfamiliar data.

In contrast, Reinforcement Learning (RL) \cite{schulman2017proximal} offers a compelling alternative for cultivating emergent reasoning by rewarding models for discovering their own logical steps rather than memorizing final answers or copying teacher CoT rationales. Indeed, a recent work \textit{\textbf{SFT Memorizes, RL Generalizes}} \cite{chu2025sft} confirms that RL-trained models often display superior generalization compared to their SFT counterparts. However, conventional RL pipelines typically depend on auxiliary neural reward models, requiring substantial resources to continuously update both policy and reward models \cite{ziegler2019fine,ouyang2022training}.
%—an especially challenging process in data-limited medical domains. 
A promising alternative, group relative policy optimization (GRPO) \cite{shao2024deepseekmath}, eliminates the need for neural reward models by employing a rule-based group-relative advantage strategy (see sec. \ref{sec:methods} for more details). This approach has demonstrated advanced reasoning, fostering capabilities while reducing computational demands in DeepSeek-R1 \cite{guo2025deepseek}. Despite its potential benefits for resource- and data-constrained domains like healthcare, GRPO remains largely unexplored in medical contexts.

In this work, we introduce \textbf{MedVLM-R1}, the first medical VLM capable of generating answers with explicit reasoning by training with GRPO for radiology VQA tasks. Our contributions are as follows:
\begin{enumerate}
    \item \textbf{Medical VLM with Explicit Reasoning}: We introduce MedVLM-R1, the first lightweight medical VLM capable of generating explicit reasoning alongside the final answer, rather than providing only the final answer.

    \item \textbf{Emerging Reasoning Without Explicit Supervision}: Unlike traditional SFT methods that require data with complex reasoning steps, MedVLM-R1 is trained using GRPO with datasets containing only final answers, demonstrating emergent reasoning capabilities without explicit supervision.

    \item \textbf{Superior Generalization and Efficiency}: MedVLM-R1 achieves robust generalization to out-of-distribution data (e.g. MRI → CT/X-ray) and outperforms larger models like Qwen2VL-72B and Huatuo-GPT-Vision-7B, despite being a compact 2B-parameter model trained on just 600 samples.
\end{enumerate}

\section{Related Work}

\paragraph{\textbf{Medical VLMs and Their Limitations.}} The rise of large-scale VLMs has spurred numerous domain-specific adaptations for healthcare, with systems such as LLaVA-Med \cite{li2023llava} and HuatuoGPT-Vision \cite{chen2024huatuogptvisioninjectingmedicalvisual} achieving impressive results in radiology VQA and related diagnostic tasks. Despite these advancements, using SFT on final-answer labels remains the dominant strategy for tailoring large models to medical domains \cite{zhang2023pmc,chen2024efficiency,zhang2024generalist,chaves2024towards}. This approach generally requires substantial amounts of high-quality image-text data (ranging from 660k \cite{li2023llava} to 32M samples \cite{wu2023towards}) which is costly to curate and often hampered by noise/privacy concerns. Moreover, the reliance on final-answer supervision provides limited scope for exposing a model’s intermediate reasoning—an important factor in building clinicians' trust. In addition, SFT-based models often overfit to narrow training distributions, leading to weaker generalization on OOD clinical scenarios.

\paragraph{\textbf{Reinforcement Learning for Enhanced Reasoning.}}
To mitigate SFT's limitations, RL \cite{silver2017mastering,christiano2017deep,ziegler2019fine,ouyang2022training,kumar2024training} has emerged as a compelling alternative for improving model interpretability and robustness. Classic RL methods, such as proximal policy optimization (PPO) \cite{schulman2017proximal}, have been widely adopted in text-based learning (\textit{e.g.,} policy shaping for LLMs) and can reward not only correctness but also the quality of intermediate reasoning steps. Recent studies suggest that while SFT “memorizes,” RL can help models “generalize” \cite{chu2025sft}, offering a more stable trajectory toward domain-transferable representations. Notably, GRPO \cite{shao2024deepseekmath} extends PPO by eliminating its (neural) value function estimator and focusing on a rule-based group-relative advantage for selecting actions, showing promise in resource-constrained settings like DeepSeek-R1 \cite{guo2025deepseek}. Such RL-driven frameworks could be particularly beneficial for medical tasks, where limited data availability, high-stakes decision-making, and the need for explicit reasoning converge. In the following section, we will detail how these insights motivate our approach.% to improving reasoning capabilities in radiology VQA.

\section{Methods}
\label{sec:methods}

\begin{figure*}[ht]
    \centering
    \begin{tikzpicture}
    \node[draw=black, 
    line width=1pt,
    rounded corners=10pt,
    text width=\textwidth,
    inner sep=5pt,
    align=left](box){\begin{wrapfigure}{r}{0.24\textwidth} % "r" for right side, and width of the wrap figure
    \vspace{-3em}
        \includegraphics[width=0.22\textwidth]{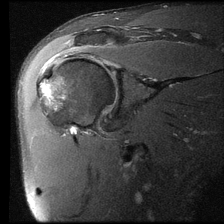}
        \end{wrapfigure}
        \small\textbf{Question Template}\vspace{0.1cm}
        
        \brown{What can be observed in this image?

        A) Cartilage degeneration  B) Labral pathology

        C) Bone fracture D) Tendonitis
        }

        \vspace{-0.25cm}
        \noindent\rule{\dimexpr0.7\textwidth}{0.4pt}
        \textbf{Model Input Prompt}\vspace{0.1cm}
        
        \vspace{-0.05cm}
        % \noindent\hdashrule{\dimexpr0.6\textwidth}{0.4pt}{1mm 0.2mm}

        \brown{\{Question\}} Your task: 

        1. Think through the question step by step, enclose your reasoning process in <think>...</think> tags. 
        
        2. Then provide the correct single-letter choice (A, B, C, D,...) inside <answer>...</answer> tags.
        
        3. No extra information or text outside of these tags.\\

    \vspace{-0.25cm}
        \noindent\rule{\dimexpr1\textwidth}{0.4pt}
        \textbf{Model Output Examples}\vspace{0.1cm}

        \blue{<think>
        The image is a grayscale MRI image of an upper arm joint. 
        The bicondylar humeral head of the humerus is visible. There is a well-defined ...
        </think>
        
        <answer>B, there is no clear indication of ... </answer>}

        \vspace{-0.25cm}
        
        \noindent\rule{\dimexpr1\textwidth}{0.4pt}
        
        \textbf{Format Reward} = 1 due to the present of all tags and no content outside the tags
        
        \textbf{Accuracy Reward} = 0.5 due to extra explanation appended after the answer 
        \vspace{-0.12cm}
    };
    \end{tikzpicture}
    \caption{\small {The template of our employed prompt, an example of model's response and reward criterion.} }
    \label{fig:main} 
\end{figure*}

\paragraph{\textbf{Overview.}} We leverage RL to incentivize explicit reasoning capabilities in medical VLMs, specifically employing GRPO due to its efficiency and effectiveness. While the seminal work \cite{shao2024deepseekmath} applies GRPO to reasoning in coding and mathematics, we adapt these principles to the medical domain, specifically radiology data (MRI, CT, X-ray). Our approach incorporates medical imaging prompts and custom reward functions designed to encourage explicit reasoning and domain-specific answer formats. This work serves as an initial exploration of using pure reinforcement learning to build a multi-modal medical reasoning model.

\paragraph{\textbf{Base Model and Prompt Template.}} We adopt a state-of-the-art VLM -- Qwen2-VL-2B \cite{wang2024qwen2} as our base model, denoted by $\pi_\theta$, where $\theta$ are the trainable parameters. Given a training dataset $\mathbf{V}$, each sample $\mathbf{v}$ consists of: 1) An image $f$, which is a radiology image and 2) a text prompt $q$, composed of the user’s question alongside a fixed system message, as illustrated in Figure~\ref{fig:main}. The VLM then produces an output $\{o\}$, which includes both a reasoning trace and a final answer in designated XML-like tags (\texttt{<think>}…\texttt{</think>} and \texttt{<answer>}…\texttt{</answer>}). Our RL objective is to optimize $\pi_\theta$ so that answers are accurate, well-formatted, and provide transparent reasoning.

\paragraph{\textbf{Group Relative Policy Optimization (GRPO).}} To encourage robust, interpretable responses, we employ GRPO~\cite{shao2024deepseekmath}, an RL algorithm that extends PPO by focusing on a group-relative advantage instead of a learned value function. Concretely, at each training step:
\begin{enumerate}
\item We sample $G$ candidate outputs $\{o_i\}_{i=1}^G$ from $\pi_{\theta_{old}}$, the model parameters before the current update.
\item We compute a reward $r_i$ for each output using a reward function (see next paragraph). Based on $r_i$ we calculate a group relative advantage $A_i$ which is normalized by the group statistics: $ A_i = \frac{r_i - {\mathrm mean(\{r_1, r_2, \cdots, r_G\})}}{{\mathrm std(\{r_1, r_2, \cdots, r_G\})}}.$ A reward above the group average is advantaged and can further incentivize the model.

\item Our VLM model $\pi_\theta$ is then updated by maximizing $\mathcal{J}_{GRPO}$ which incorporates a clipped regularization on the relative advantage estimation for model preference alignment and training stability:

\begin{equation}
\begin{split}
    & \mathcal{J}_{GRPO}(\theta) = \mathbb{E}_{\mathbf{v} \sim P(\mathbf{V})}\mathbb{E}_{\{o_i\}_{i=1}^G \sim \pi_{\theta_{old}}(\cdot|\mathbf{v})}  \\
    & \frac{1}{G}\sum_{i=1}^G \Bigl[\min \Bigl( r_{i}^{\text{ratio}} A_i, \text{clip} \left( r_{i}^{\text{ratio}}, 1 \pm \epsilon \right)  A_i\Bigl) - \beta \mathbb{D}_{KL}\left(\pi_{\theta} || \pi_{ref}\right)\Bigl]
\end{split}
\label{eq:GRPO}
\end{equation}
with $r_{i}^{\text{ratio}} = \frac{\pi_\theta(o_i|\mathbf{v})}{\pi_{\theta_{\text{old}}}(o_i|\mathbf{v})}$. An additional \textit{Kullback–Leibler} term $\mathbb{D}_{\text{KL}}(\pi_{\theta} \Vert \pi_{ref})$ is applied to penalize divergence from a reference model $\pi_{\text{ref}}$ (the initial checkpoint), helping prevent catastrophic forgetting.  $\epsilon, \beta \in \mathbb{R}\geq0$ control the regularization strengths.
\end{enumerate}

\paragraph{\textbf{Reward function.}} Specifically, for multiple-choice medical VQA tasks, we propose a two-part rule-based reward function, inspired by \cite{guo2025deepseek}:

\noindent 1) \textit{Format Reward.} We incentivize outputs that provide a reasoning trace within the tags \texttt{<think>} ... \texttt{</think>} and a succinct final answer within the tag \texttt{<answer>} ... \texttt{</answer>}. If all four tags are present exactly once and no content is present outside these tags, we assign a format reward of 1. Any missing/duplicated tags or content outside yield 0.

\noindent 2) \textit{Accuracy Reward.} After verifying the correct format, we evaluate the correctness of the final answer. Specifically, If the letter choice \texttt{A, B, C, D,}\ldots inside the \texttt{<answer>} ... \texttt{</answer>} tag and it responds with the ground-truth choice exactly, that is an exact match with a reward of 1 point. Further, if the letter is correct but contains additional explanations or uses the choice content instead of the corresponding letter (e.g., “A: Pulmonary nodule” or "Pulmonary nodule"), that is a partial match with a 0.5 points reward. However, if the letter does not match, is missing, or the answer is not enclosed in the answer tag, that is an incorrect or missing answer and no reward would be granted (0 points).

The total reward $r_i \in [0, 2]$ is the sum of both format and accuracy reward. By structuring the reward function in this hierarchy (format before correctness), we guide the model to first adopt the desired response structure, then refine its answer selection for accurate, interpretable medical reasoning. It is worth noting that both terms are necessary since without \textit{Format Reward}, the final answer cannot be extracted while without \textit{Accuracy Reward}, the model cannot converge.

\section{Experiments}

\paragraph{\textbf{Dataset.}}
We conduct our experiments using the HuatuoGPT-Vision \cite{chen2024huatuogptvisioninjectingmedicalvisual} evaluation dataset, which is a processed and combined dataset from several publicly available medical VQA benchmarks, including VQA-RAD \cite{lau2018dataset}, SLAKE \cite{liu2021slake}, PathVQA \cite{he2020pathvqa}, OmniMedVQA \cite{hu2024omnimedvqa}, and PMC-VQA \cite{zhang2023pmc}. In total, the dataset comprises 17,300 multiple-choice questions linked to images covering various medical imaging modalities, with 2–6 possible choices per question. For this study, we focus on radiology modalities: CT, MRI, and X-ray. Specifically, we use 600 MRI image-question pairs for training and set aside 300 MRI, 300 CT, and 300 X-ray pairs for testing. The MRI test set is used as in-domain test, whereas the CT and X-ray test set serve as OOD test.

\paragraph{\textbf{Implementation details.}}
We adopt Qwen2-VL-2B as our base VLM. This model is originally trained on data from curated web pages, open-source datasets, and synthetic sources. To adapt it to the medical domain, we employ the GRPO reinforcement learning framework outlined in Section~\ref{sec:methods}. Our implementation builds on the public VLM reasoning repositories \cite{open-r1-multimodal,chen2025r1v,shen2025vlmr1}. We perform fine-tuning on two NVIDIA A100 SXM4 80GB for 300 steps, using a batch size of 2, which takes approximately 4 hours. Generation candidate number $G$ is set to 6. The other training optimization hyper-parameters are set as suggested by \cite{chen2025r1v}.

\paragraph{\textbf{Baseline methods and evaluation metric.}}
We compare MedVLM-R1 with the following baselines: 1. Qwen2-VL family \cite{Qwen-VL} including Qwen2-VL-2B (the unmodified base model), Qwen2-VL-7B and -72B which are the large/huge model variants. 2. HuatuoGPT-vision \cite{chen2024huatuogptvisioninjectingmedicalvisual}: A medical VLM built upon Qwen2-VL-7B. 3. SFT: The same Qwen2-VL-2B base model fine-tuned with standard SFT, using the same training setting with 600 MRI question-answer pairs. We apply negative log-likelihood as the loss function to carry out the SFT training. All baselines use a simple prompting format, \textit{e.g.,} \texttt{\{Question\} Your task: provide the correct single-letter choice (A, B, C, D, …).} In contrast, MedVLM-R1 uses the RL-based prompt as described in Section~\ref{sec:methods}, designed to elicit explicit reasoning. For evaluation, each model receives one point for the correct single-letter answer and zero otherwise. In the test of MedVLM-R1, only the correct choice enclosed in the \texttt{<answer>...</answer>} tag is scored as correct; any deviation from this format, even if semantically correct, results in a zero score.

\section{Results and Discussion}

\paragraph{\textbf{Overall Performance.}}

\begin{figure*}[!ht]
    \centering
    \begin{tikzpicture}[node distance=1cm] % Add node distance parameter
    
    % First node (original)
    \node[draw=black, 
    line width=1pt,
    rounded corners=10pt,
    text width=\textwidth,
    inner sep=5pt,
    align=left](box1){ % Changed name to box1
        \begin{wrapfigure}{r}{0.22\textwidth}
        \vspace{-2.5em}
            \includegraphics[width=0.22\textwidth]{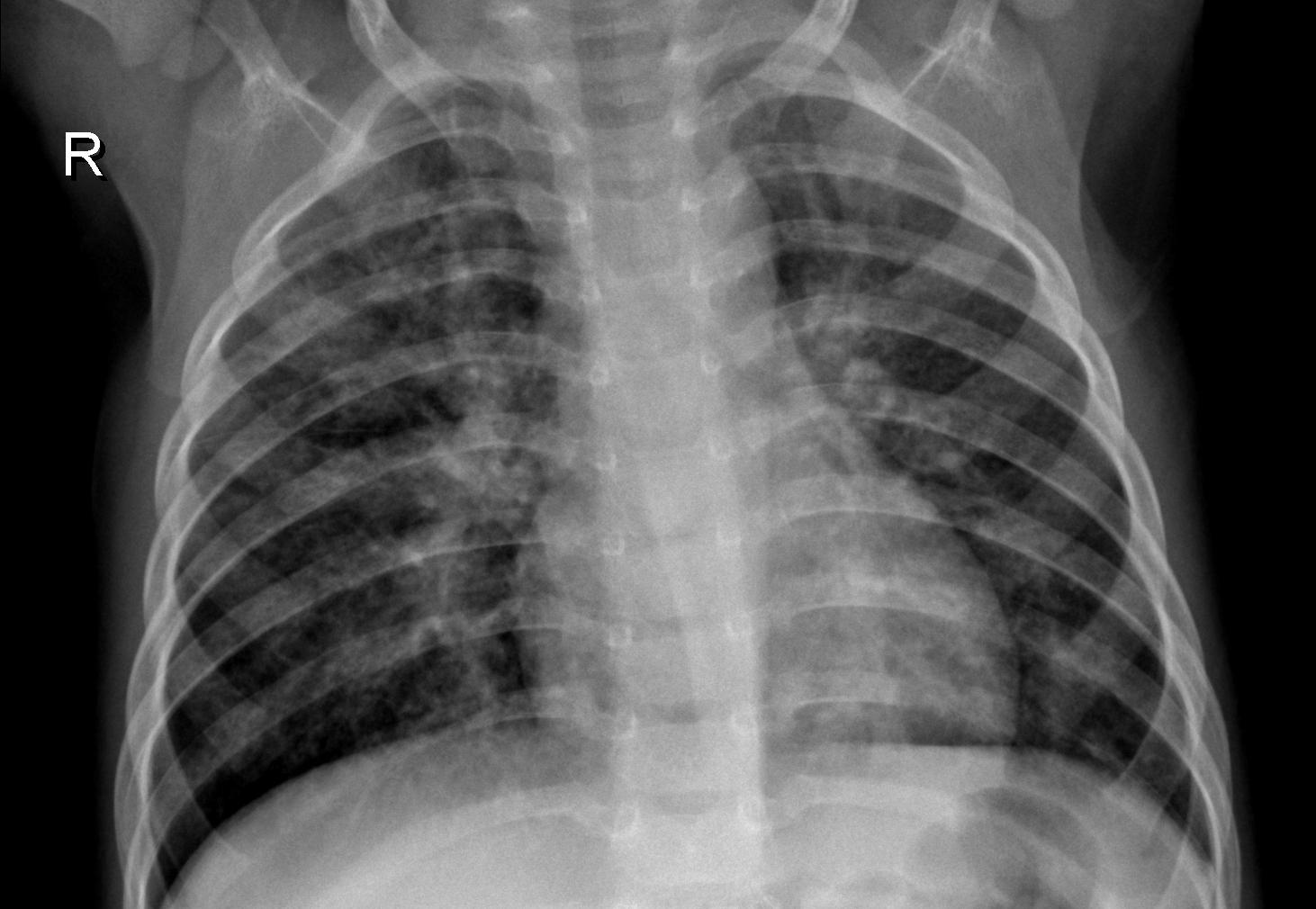}
        \end{wrapfigure}
        \vspace{0.1cm}
        \scriptsize{\brown{Problem 1: What can be observed in this image?

        A) Lungs  B) Bladder C) Brain D) Heart
        }}
        
        \vspace{0.1cm}

        \scriptsize{\textbf{MedVLM-R1 Output:}\vspace{0.1cm}

        \blue{<think>
        
The image is a chest X-ray, which is a type of radiographic image used to visualize the internal structures of the body, particularly the lungs and bones. The presence of lung markings and the ribcage are characteristic features of a chest X-ray.

        </think>
        \vspace{0.06cm}
        
        <answer>A</answer>}
        % \noindent\rule{\dimexpr1\textwidth}{0.4pt}
        \scriptsize{\textbf{\green{Groundtruth Answer: A}}}
        % \vspace{0.2cm}
        }
    };
    
    % Second node (copy of the first)
    \node[draw=black, 
    line width=1pt,
    rounded corners=10pt,
    text width=\textwidth,
    inner sep=5pt,
    align=left,
    below=0.1cm of box1](box2){ % Changed name to box1
        \begin{wrapfigure}{r}{0.22\textwidth}
        \vspace{-2.5em}
            \includegraphics[width=0.22\textwidth]{figs/mdb146.png}
        \end{wrapfigure}
        \vspace{0.1cm}
        \scriptsize{\brown{Problem 2: What content appears in this image?
        
A) Cardiac tissue, B) Breast tissue, C) Liver tissue, D) Skin tissue

        }}
        
        \vspace{0.1cm}

        \scriptsize{\textbf{MedVLM-R1 Output:}\vspace{0.1cm}

        \blue{<think>
            The image appears to be a mammogram, which is a type of X-ray used to detect abnormalities in the breast tissue. The image shows the breast tissue with various densities and patterns, which are typical of mammograms.
            </think>

        \vspace{0.06cm}
        
       <answer>B</answer>
       }
        % \noindent\rule{\dimexpr1\textwidth}{0.4pt}
        \scriptsize{\textbf{\green{Groundtruth Answer: B}}}
        % \vspace{0.2cm}
        }
    };
        \node[draw=black, 
    line width=1pt,
    rounded corners=10pt,
    text width=\textwidth,
    inner sep=5pt,
    align=left,
    below=0.1cm of box2](box3){ % Changed name to box1
        \begin{wrapfigure}{r}{0.22\textwidth}
        \vspace{-2.5em}
            \includegraphics[width=0.22\textwidth]{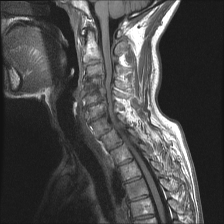}
        \end{wrapfigure}
        \vspace{0.1cm}
        \scriptsize{\brown{Problem 3: What is the finding in this image?
        
A) Stomach ulcer, B) Nerve compression, 

C) Cord pathology, D) Liver cirrhosis

        }}
        
        \vspace{0.1cm}

        \scriptsize{\textbf{MedVLM-R1 Output:}\vspace{0.1cm}

        \blue{<think>
To determine the correct answer, we need to identify the finding in the image provided. The MRI scan in question is likely a view of a spine, not the stomach, or the liver. Stomach ulcers, liver cirrhosis, and spinal cord pathology can be diagnosed with MRI, but the image in question shows a spinal view, not a gastrointestinal or abdominal view. Therefore, the only reasonable options left are nerve compression and cord pathology.
</think>

        \vspace{0.06cm}
        
       <answer>C</answer>
       }
        % \noindent\rule{\dimexpr1\textwidth}{0.4pt}
        \scriptsize{\textbf{\green{Groundtruth Answer: C}}}
        % \vspace{0.2cm}
        }
    };
    \node[draw=black, 
    line width=1pt,
    rounded corners=10pt,
    text width=\textwidth,
    inner sep=5pt,
    align=left,
    below=0.1cm of box3](box4){ % Changed name to box1
        \begin{wrapfigure}{r}{0.22\textwidth}
        \vspace{-2.5em}
            \includegraphics[width=0.22\textwidth]{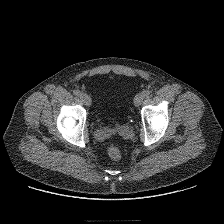}
        \end{wrapfigure}
        \vspace{0.1cm}
        \scriptsize{\brown{Problem 4: What is the specific abnormality present in this image?

A) Gastritis, B) Arthritis, C) Appendicitis, D) Urolithiasis

        }}
        
        \vspace{0.1cm}

        \scriptsize{\textbf{MedVLM-R1 Output:}\vspace{0.1cm}

        \blue{<think> The image presented is a transverse CT scan of the abdomen and pelvis. The presence of calculi (urines filled with stones or grit) in the pelvic organs is a consistent finding in urolithiasis. </think>

        \vspace{0.1cm}
        
       <answer>D</answer>
       }
        % \noindent\rule{\dimexpr1\textwidth}{0.4pt}
        \scriptsize{\textbf{\green{Groundtruth Answer: D}}}
        % \vspace{0.2cm}
        }
    };
%         \node[draw=black, 
%     line width=1pt,
%     rounded corners=10pt,
%     text width=\textwidth,
%     inner sep=5pt,
%     align=left,
%     below=0.1cm of box3](box4){ % Changed name to box1
%         \begin{wrapfigure}{r}{0.22\textwidth}
%         \vspace{-2.5em}
%             \includegraphics[width=0.22\textwidth]{figs/brain001109.png}
%         \end{wrapfigure}
%         \vspace{0.1cm}
%         \scriptsize{\brown{Problem 4: What can be observed in this image?
        
% A) Abscess formation, B) Fractured bone, C) Normal tissue, D) Chronic infarct

%         }}
        
%         \vspace{0.15cm}

%         \scriptsize{\textbf{MedVLM-R1 Output:}\vspace{0.1cm}

%         \blue{<think> To answer this question, we need to identify any abnormalities or structural changes within the MRI scan. The MRI scan in question displays a section of the brain, with the left-sided lower part appearing different from the right side. The left side shows irregular and more expansive regions compared to the right side. Additionally, the contrast in tissue density seems to be more visible on the left side, with darker regions indicating different tissue densities or possibly abnormal regions.

% </think>

%         \vspace{0.1cm}
        
%        <answer>D</answer>
%        }
    
%         % \noindent\rule{\dimexpr1\textwidth}{0.4pt}
%         \scriptsize{\textbf{\green{Groundtruth Answer: D}}}
%         % \vspace{0.2cm}
%         }
%     };
    \end{tikzpicture}
 \caption{\small{Medical VQA examples of MedVLM-R1 on X-ray (1, 2), MRI (3) and CT (4).}}
    \label{fig:vqa} 
\end{figure*}

Table~\ref{tab:prediction} summarizes both in-domain (ID) and out-of-domain (OOD) performance for various VLMs. Note that ID/OOD comparisons specifically refer to models fine-tuned on MRI data. Unsurprisingly, VLMs fine-tuned with both GRPO and SFT significantly outperform zero-shot general-purpose VLMs on in-domain tasks. Our GRPO-trained model shows very strong OOD performance, achieving a \textbf{16\%} improvement on CT and a \textbf{35\%} improvement on X-ray compared to SFT counterparts, underscoring GRPO's superior generalizability. Furthermore, despite being a compact 2B-parameter model trained on just 600 samples, MedVLM-R1 outperforms larger models like Qwen2-VL-72B and HuatuoGPT-Vision-7B, with the latter being specifically trained on large-scale medical data. This highlights the immense potential of RL-based training methods for efficient and scalable medical VLM development.

\paragraph{\textbf{Reasoning Competence and Interpretability.}} Beyond strong generalization, a central strength of MedVLM-R1 is its ability to produce explicit reasoning—a capability absent in all baselines. As illustrated in Figure~\ref{fig:vqa}, MedVLM-R1 presents a logical thought process within the \texttt{<think>} tag, with the final decision enclosed in the \texttt{<answer>} tag. Notably, for relatively simpler questions (problem 1 and 2), the reasoning appears cogent and aligned with medical knowledge. However, more complex queries sometimes reveal heuristic or just partial reasoning. For example, in the third sample, the model arrives at the correct answer via the \textbf{process of elimination} rather than detailed medical analysis, suggesting it leverages cue-based reasoning instead of domain expertise. Likewise, in some instances (e.g., question 4), the causal chain between reasoning and conclusion remains unclear, raising the question of whether the model merely \textbf{retrofits an explanation after predicting the correct answer}. Despite these imperfections, MedVLM-R1 represents a notable step toward interpretability in radiological decision-making.

\begin{table}[!t]
\centering
\setlength{\tabcolsep}{1mm}{}
\caption{\small {Results of VQA-VLMs on MRI (in-domain), and CT and X-Ray (out-of-domain) modalities. "$-2$B" indicates the model has $2$ billion parameters, etc.}}
\label{tab:prediction}
\scalebox{0.85}{
\begin{tabular}{l|c|ccc|c}
\toprule
\multirow{2}{*}{Method} & \multirow{2}{*}{\shortstack{Num. of Seen \\ Medical Sample}} & \multicolumn{3}{c|}{In-Domain / Out-of-Domain} & \multirow{2}{*}{Average} \\ 
\cline{3-5}
         &  & (MRI$\rightarrow$MRI) & (MRI$\rightarrow$CT) & (MRI$\rightarrow$X-ray) & \\ \midrule
Random Guess & / & $25.00$ & $30.25$ & $26.00$ & $27.08$   \\ 
\hline
\multicolumn{6}{c}{\footnotesize\textit{Zero-shot VLM}} \\
\hline
Qwen2-VL-2B & / & $61.67$ & $50.67$ & $53.00$ & $55.11$   \\ 
Qwen2-VL-7B & / & $72.33$ & $68.67$ & $66.63$ & $69.21$   \\ 
Qwen2-VL-72B & / & $68.67$ & $60.67$ & $72.33$ & $67.22$   \\ 
\hline
\multicolumn{6}{c}{\footnotesize\textit{Zero-shot Medical VLM}} \\
\hline
Huatuo-GPT-vision-7B & 1,294,062 & $71.00$ & $63.00$ & $\mathbf{73.66}$ & $69.22$   \\ 
\hline
\multicolumn{6}{c}{\footnotesize\textit{MRI fine-tuned VLM}} \\
\hline
Qwen2-VL-2B (SFT) & 600 & $94.00$ & $54.33$ & $34.00$ & $59.44$   \\ 
\textbf{Ours-2B (GRPO)} & 600 & $\mathbf{95.33}$ & $\mathbf{70.33}$ & $69.00$ & $\mathbf{78.22}$   \\
\bottomrule
\end{tabular}
}
\end{table}

\paragraph{\textbf{Limitations.}} Although MedVLM-R1 demonstrates promising results in MRI, CT, and X-ray datasets, several limitations remain: 1. Modality Gaps: When tested on other medical modalities (e.g., pathology or OCT images), the model fails to converge. We hypothesize this arises from the base model’s insufficient exposure to such modalities during pre-training. 2. Closed-Set Dependence: The current approach is tailored to multiple-choice (closed-set) VQA. In open-ended question settings where no predefined options are provided, the model’s performance degrades substantially. This is also a common challenge for many VLMs. 3. Superficial/hallucinated Reasoning: In some reasoning cases, MedVLM-R1 provides a correct answer without offering a meaningful reasoning process (e.g., \texttt{<think>To determine the correct observation from this spine MRI, let's analyze the image.</think><answer>B</answer>}). Moreover, sometimes the model concludes a correct choice while providing an inference that can lead to another answer. This phenomenon underscores that even models designed for explainability can occasionally revert to superficial/hallucinated justifications, highlighting an ongoing challenge in generating consistently transparent and logically sound rationales. Regarding all these issues, we believe the current 2B-parameter scale of our base model constitutes a potential bottleneck, and we plan to evaluate MedVLM-R1 on larger VLM backbones to address these concerns.

\section{Conclusion}

We present MedVLM-R1, a medical VLM that integrates GRPO-based reinforcement learning to bridge the gap between accuracy, interpretability, and robust performance in radiology VQA. By focusing on explicit reasoning, the model fosters transparency and trustworthiness—qualities essential in high-stakes clinical environments. Our results demonstrate that RL-based approaches generalize better than purely SFT methods, particularly under OOD settings. Although VLM-based medical reasoning is still at a nascent stage and faces considerable challenges, we believe that its potential for delivering safer, more transparent AI-driven healthcare solutions will be appreciated and should be encouraged.

\section{Acknowledgements}
This work is partially funded by the European Research Council (ERC) project Deep4MI (884622). Mr. Wu is supported by the Engineering and Physical Sciences Research Council (EPSRC) under grant EP/S024093/1 and GE HealthCare. Mr. Liu is supported by the Clarendon Fund. Ms. Zhu is supported by the Engineering and Physical Sciences Research Council (EPSRC) under grant EP/S024093/1 and Global Health R\&D of the healthcare business of Merck KGaA, Darmstadt, Germany, Ares Trading S.A. (an affiliate of Merck KGaA, Darmstadt, Germany), Eysins, Switzerland (Crossref Funder ID: 10.13039 / 100009945). Dr. Li is supported by a Postdoc Mobility Grant from SNSF. Dr. Chen is funded by Royal Society (RGS/R2/242355). Dr. Ouyang is supported by UKRI grant EP/X040186/1.

% ---- Bibliography ----
%
\bibliographystyle{splncs04}
\bibliography{refs}

\end{document}